\documentclass[conference]{IEEEtran}
\IEEEoverridecommandlockouts
\usepackage{cite}
\usepackage{amsmath,amssymb,amsfonts}
\usepackage{algorithmic}
\usepackage{graphicx}
\usepackage{textcomp}
\usepackage{xcolor}
\newcommand{\tabincell}[2]{\begin{tabular}{@{}#1@{}}#2\end{tabular}}
\def\BibTeX{{\rm B\kern-.05em{\sc i\kern-.025em b}\kern-.08em
    T\kern-.1667em\lower.7ex\hbox{E}\kern-.125emX}}
\begin{document}

\title{LighTN: Light-weight Transformer Network for Performance-overhead Tradeoff in Point Cloud Downsampling\\
\thanks{This work was supported by the National Natural Science Foundation of China under grant No.61972030.}
}

\author{\IEEEauthorblockN{1\textsuperscript{st} Xu Wang}
\IEEEauthorblockA{\textit{School of Computer and Information} \\
\textit{Technology, Beijing Jiaotong University}\\
Beijing, China \\
xu.wang@bjtu.edu.cn}
\and
\IEEEauthorblockN{2\textsuperscript{nd} Yi Jin*}
\IEEEauthorblockA{\textit{School of Computer and Information} \\
\textit{Technology, Beijing Jiaotong University}\\
Beijing, China \\
yjin@bjtu.edu.cn}
\and
\IEEEauthorblockN{3\textsuperscript{rd} Yigang Cen}
\IEEEauthorblockA{\textit{School of Computer and Information} \\
\textit{Technology, Beijing Jiaotong University}\\
Beijing, China \\
ygcen@bjtu.edu.cn}
\and
\IEEEauthorblockN{4\textsuperscript{th} Tao Wang}
\IEEEauthorblockA{\textit{School of Computer and Information} \\
\textit{Technology, Beijing Jiaotong University}\\
Beijing, China \\
twang@bjtu.edu.cn}
\and
\IEEEauthorblockN{5\textsuperscript{th} Bowen Tang}
\IEEEauthorblockA{\textit{Computer Architecture, Institute of} \\
\textit{Computing Technology Chinese Academy}\\
\textit{of Sciences} Beijing, China \\
tangbowen@ict.ac.cn}
\and
\IEEEauthorblockN{6\textsuperscript{th} Yidong Li}
\IEEEauthorblockA{\textit{School of Computer and Information} \\
\textit{Technology, Beijing Jiaotong University}\\
Beijing, China \\
ydli@bjtu.edu.cn}
}
\maketitle

\begin{abstract}
Compared with traditional task-irrelevant downsampling methods, task-oriented neural networks have shown improved performance in point cloud downsampling range. Recently, Transformer family of networks has shown a more powerful learning capacity in visual tasks. However, Transformer-based architectures potentially consume too many resources which are usually worthless for low overhead task networks in downsampling range. This paper proposes a novel light-weight Transformer network (LighTN) for task-oriented point cloud downsampling, as an end-to-end and plug-and-play solution. In LighTN, a single-head self-correlation module is presented to extract refined global contextual features, where three projection matrices are simultaneously eliminated to save resource overhead, and the output of symmetric matrix satisfies the permutation invariant. Then, we design a novel downsampling loss function to guide LighTN focuses on critical point cloud regions with more uniform distribution and prominent points coverage. Furthermore, We introduce a feed-forward network scaling mechanism to enhance the learnable capacity of LighTN according to the expand-reduce strategy. The result of extensive experiments on classification and registration tasks demonstrates LighTN can achieve state-of-the-art performance with limited resource overhead.
\end{abstract}


\section{Introduction}
Simplification of point sets in an original point cloud input, referred to as downsampling, is a fundamental work in perception of 3D visual scenes with applications in many intelligent systems, such as autonomous driving, assistive robots, and digital city. For example, it is capable of
\begin{figure}[!hbt]
    \centering
    \includegraphics[scale=0.35]{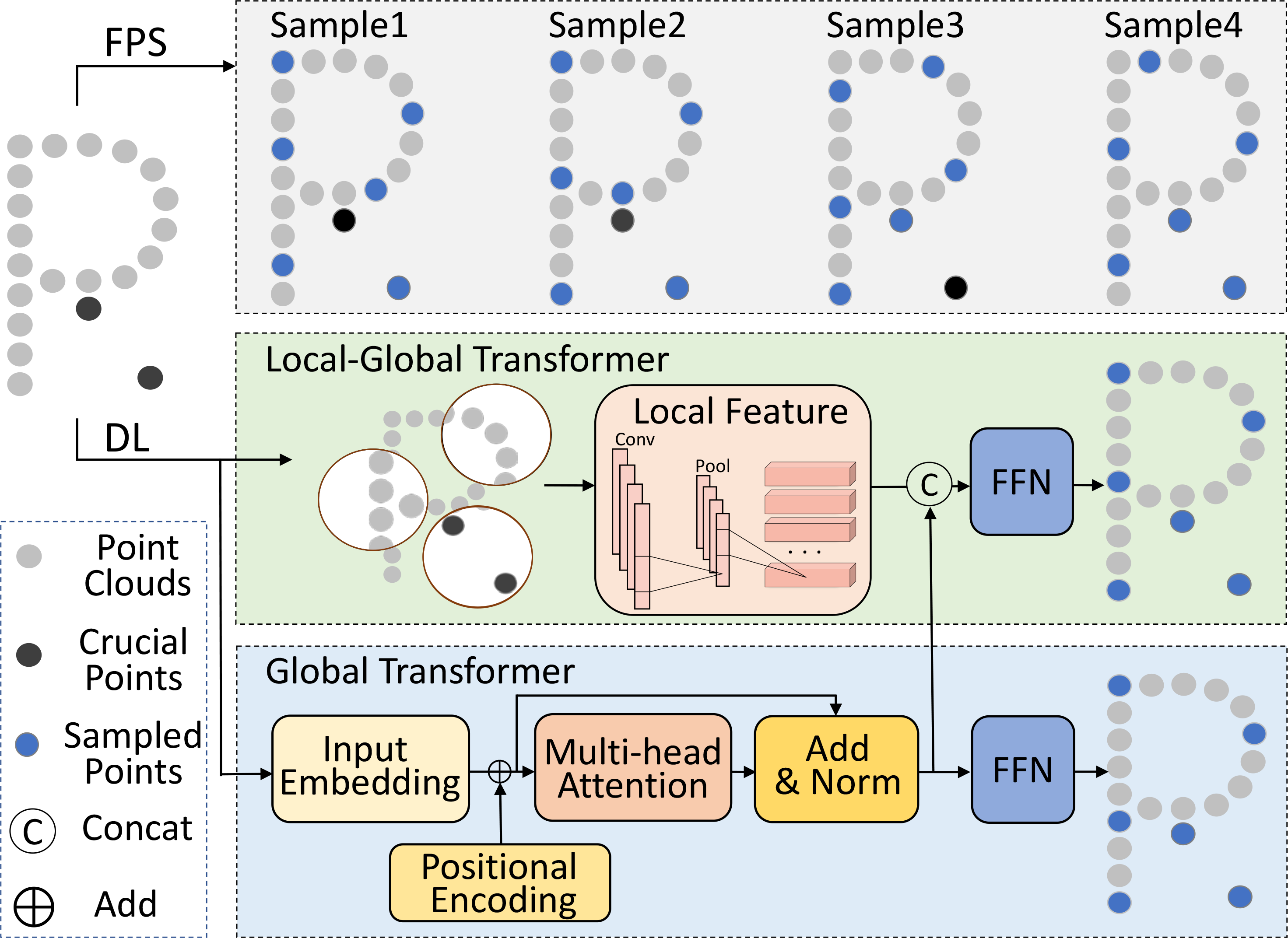}
    \caption{Visualization of point cloud downsampling methods by FPS, local-global-based Transformer and global-based Transformer. The black dots in the input is used as critical points to distinguish the letter \textbf{R} and \textbf{P}. DL means deep learning and Norm means normalization.}
    \label{fig1}
\end{figure}
reducing the computation load, storage space, and communication bandwidth on low-power devices or terminals when facing tens of thousands of points with limited resource overhead. In practical scenarios, the capacity of point set is not always necessarily proportional to point cloud quality and recognition effect, especially with large-scale or dense points. However, point cloud downsampling remains a challenging task since the downstream process expects to take critical points with complete context information.

The traditional two-stage downsampling approaches have been applied in many works, e.g., farthest point sampling (FPS) \cite{li2021point} and random sampling (RS) \cite{hu2020randla}. However, the sampled points are based on low-level information irrelevant to downstream tasks, without considering deep semantic features and task-related information. Moreover, traditional downsampling strategy may potentially consume more resource overhead for desired precision that may even exceed the resource savings brought by downsampling, as shown in the top part of Figure \ref{fig1}.

Recently, many works have concentrated on deep learning (DL) to identify critical points by embedding the point cloud into high-dimensional feature space. There are two main lines of research in point cloud downsampling range to design a deep model for feature learning. The first research line directly integrates the designed downsampling layer into a specific-defined single-task neural network. In this line, both global \cite{Nezhadarya_2020_CVPR} and local-global \cite{wu20213d, DBLP:conf/aaai/Wu0WL20} features are used to find the critical points in point cloud input. Although these approaches successfully simplified the point cloud, the specific-defined network architecture can not ensure universality to other tasks. Moreover, integrated design is negative to priori defined network architectures and pre-trained parameters for two reasons: \textit{1)} a priori defined network with complex framework and high accuracy will cost expensive resources on training, and \textit{2)} any slight variation of network architectures may receive unpredictable changes in performance. In view of these shortcomings, we are interested in accomplishing a universal downsampling design without changing existing priori defined task network structures and pre-trained parameters.

Fortunately, the second research line is to design an independent downsampling block that subjects to a subsequent pre-trained task network. This idea was first developed by Dovrat $et\;al.$ \cite{dovrat2019learning}, called S-NET, a downsampling block separate from the task network. S-NET shows better performance on critical point extraction than traditional approaches. Following this work, Lang $et\;al.$ \cite{lang2020samplenet} and Qian $et\;al.$ \cite{qian2020mops} utilize soft projection strategy to promote the learned points to be the proper subset of original point cloud for performance improvement. However, the backbone of S-NET, SampleNet \cite{lang2020samplenet} and MOPS-Net \cite{qian2020mops} for high-dimensional feature extraction is solely based on PointNet \cite{qi2017pointnet}, which has limited expression and learning ability. Lately, Transformer has shown great strength in dealing with disordered and unstructured point cloud data. The potential Transformer-based downsampling frameworks are outlined in Figure \ref{fig1}. However, it is hard to apply Transformer-based architecture directly to replace PointNet, because existing Transformers subsumes too many resources, which may exceed the resource saving brought by downsampling. To our knowledge, above challenges have not been overcome elegantly yet in downsampling range.

To solve the above problems, this paper presents a novel end-to-end and light-weight Transformer network, LighTN, for task-oriented point cloud downsampling. The core component of LighTN is a single-head self-correlation module, which can calculate the attention score without matrices projection of Query, Key and Value. It is worth noting that the self-correlation mechanism is more suitable for calculating the attention score because its internal symmetry matrix satisfies the permutation invariance of point cloud. Next, the point-wise global features weighted by attention scores pave the way for the subsequent Feed-forward Network (FFN) to produce an optimal subset of point cloud input in specific task applications. Besides, to alleviate the decline of learnable parameters caused by single-head self-correlation architecture, this paper expands the linear layer in FFN to increase the depth and learnable parameters for LighTN by expand-reduce strategy. With this operation, LighTN can be developed to achieve optimal critical points extraction with controlled resource overhead. Highlight that, given a priori defined task network with a specific objective function (task loss), LighTN can automatically find the proper subset of point cloud with optimal performance without any changes to the task network with pre-trained parameters.

The key contributions of this work are summarized as follows:
\begin{itemize}
    \item To handle the challenges in point cloud downsampling with limited resource overhead, we propose a plug-and-play and light-weight Transformer framework, named LighTN, as a task-oriented and end-to-end solution.
    \item We design a novel sampling loss function to promote more uniform distribution and prominent points coverage of sampled point clouds.
    \item Extensive experiments on classification and registration tasks demonstrate LighTN achieves improved performance and tolerable resource overhead for point cloud downsampling.
\end{itemize}

The structure of this paper is as follows. In section \ref{sec: related work}, we introduce short overviews of deep learning on point clouds and downsampling methods. Then, we describe the architecture of LighTN both in outline and detail in section \ref{sec: method}. In section \ref{sec: experiments}, we validate the performance of LighTN in classification and registration tasks using standard benchmark datasets and compare the result with state-of-the-art models. Finally, we present the conclusion in section \ref{sec: conclusion}.
\begin{figure*}[!hbt]
    \centering
    \includegraphics[scale=0.6]{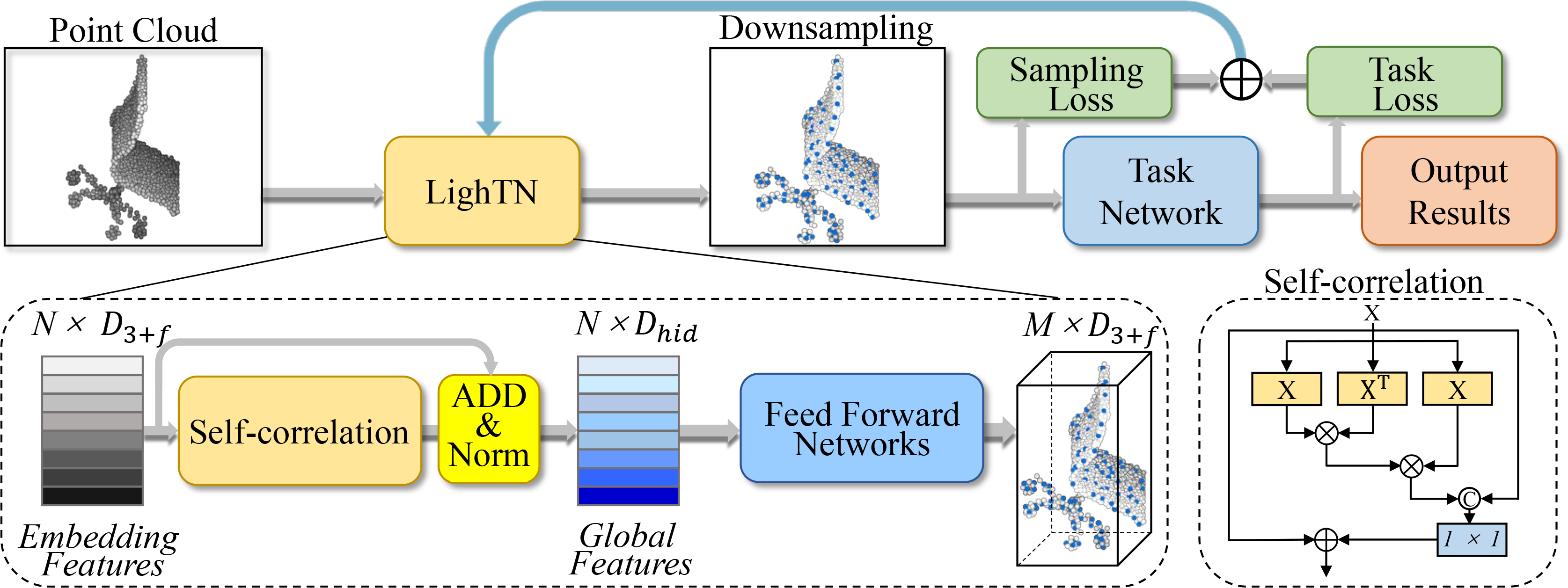}
    \caption{The framework of point cloud downsampling network. The whole network consists of a task-oriented Light-weight Transformer Network (LighTN) and a downstream task network. The point cloud input $P$ with $N$ points is trained on LighTN and then output simplified subset $Q$ with $M$ points that paves the way for downstream task network. Notable, task network is pre-trained where the weight parameters are kept fixed in training and testing stage of LighTN. Training loss includes sampling loss and task loss, which jointly optimize the weight parameters of LighTN.}
    \label{fig2}
\end{figure*}
\section{Related Work}
\label{sec: related work}
\subsection{Deep Learning for 3D Point Clouds}
Unlike structured 2D images, point cloud comprises three-dimensional coordinates of unordered and irregular points, making the powerful 2D convolutional network unusable directly. Several types of research mainly focus on transforming 3D point clouds into regular representation in terms of multi-views \cite{DBLP:conf/eccv/ZhouZLSZZFQ18, DBLP:conf/cvpr/LiZFTT20} and voxel grids \cite{DBLP:conf/corl/ZhouSZAGOGNV19, DBLP:conf/cvpr/GojcicZWW19} to solve above problem. Zhou $et\;al.$ \cite{DBLP:conf/eccv/ZhouZLSZZFQ18} proposed a multi-view descriptor, MVDesc, to learn local features from each patch of view. They collect images from 3 fixed views for each object to enhance efficiency. In order to discard task-independent fixed viewpoints, Li $et\;al.$ \cite{DBLP:conf/cvpr/LiZFTT20} present a 2D convolution framework based on variable views to extract local view-based features by introducing differentiable renderers. Their experiments showed that under eight viewpoints could obtain saturated performance. In contrast with multi-view method based on 2D convolution, voxel grid approaches use 3D convolution to achieve specific tasks. Maturana $et\;al.$ \cite{DBLP:conf/iros/MaturanaS15} voxelized the point cloud objects using occupancy grids, and then leveraged the standard 3D convolutional neural network to extract features from the raw volumetric data. Furthermore, Zhou $et\;al.$ \cite{DBLP:conf/corl/ZhouSZAGOGNV19} utilize vertical column voxelization to improve computational efficiency. Although regular representation methods have been applied in many works, they destroy the spatial structure of point cloud with inherent permutation invariant.

On the other hand, PointNet \cite{qi2017pointnet} is the pioneering work that operates directly on point cloud by using symmetric function (Max Pooling) to ensure permutation invariant characteristics. However, limited by the learning capacity of PointNet, there is still room to improve the performance of feature extraction. Some methods \cite{DBLP:conf/nips/QiYSG17, DBLP:conf/cvpr/HuangWN18, yan2020pointasnl, DBLP:conf/cvpr/SheshappanavarK20} have been proposed to extend PointNet by combining local-global geometric information for higher performance. Recently, Transformer family of networks has shown more powerful learning capacity in visual tasks, especially the output features after going through the self-attention mechanism containing refined global context information \cite{wang2021attention}. The work of three early researches \cite{DBLP:journals/cvm/GuoCLMMH21, DBLP:journals/access/EngelBD21, DBLP:journals/corr/abs-2012-09164} demonstrated the Transformer-based approaches possess inherently permutation invariant for point cloud data. For example, Guo $et\;al.$ \cite{DBLP:journals/cvm/GuoCLMMH21} propose a local-global Transformer, named PCT, with 2 layer local neighbor embedding \cite{DBLP:journals/tog/WangSLSBS19} and 4 stacked offset-attention blocks. It should be pointed out that compared with PointNet, the classification accuracy of PCT is increased by 4\%, but the computational cost is increased by more than 5 times. Besides, the work of Point Transformer \cite{DBLP:journals/access/EngelBD21} exposed that the network size of Point Transformer (51MB) is almost 5.4 times than PointNet (9.4MB) after 3.6\% improvement in classification accuracy. Unfortunately, above Transformer-based architectures potentially consume too many resources which makes the Transformer usually worthless for low overhead task networks in downsampling range. The recent work in \cite{DBLP:conf/iclr/MehtaGIZH21} addresses the problem of model size by designing singe-head attention and light-weight FFN in machine translation and language modeling tasks. Inspired by this work, we present a light-weight Transformer that captures the refined global context information with limitied computation load and storage space, the details can be seen in section \ref{sec: Light-weight Transformer}.

\subsection{Point Cloud Downsampling Methods}
In the early works, non-learned predetermined point cloud downsampling approaches have been widely adopted. For example, FPS is an important downsampling approach frequently used in many point-based networks, e.g., PointNet++ \cite{DBLP:conf/nips/QiYSG17}, PointCNN \cite{DBLP:conf/nips/LiBSWDC18}, 41-spec-cp \cite{DBLP:conf/eccv/WangSS18} and RS-CNN \cite{DBLP:conf/cvpr/LiuFXP19}. Besides, RS is also a crucial algorithm to process large-scale point clouds in deep learning scope with excellent computational efficiency, including the works of VoxelNet \cite{DBLP:conf/cvpr/ZhouT18}, MeteorNet \cite{DBLP:conf/iccv/LiuYB19}, RandLA-Net \cite{DBLP:conf/cvpr/Hu0XRGWTM20} and P2B \cite{DBLP:conf/cvpr/QiFCZX20}. However, non-learned downsampling methods are task-irrelevant, which cannot optimize the data distribution of sampled points during model training. Therefore, with the continuous increase of downsampling ratio, the output precise of following task network decreases rapidly.

Later on, the learning-based approaches show superior performance on point cloud downsampling. Noticeably, as an emerging downsampling field that deals with priori defined task networks with/without pre-trained weights, there is not a vast array of works as exemplary. The pioneering research of Dovrat $et\;al.$ \cite{dovrat2019learning} presented S-NET, the first task-oriented point cloud downsampling network. Following this work, Lang $et\;al.$ \cite{lang2020samplenet} propose a soft projection strategy to alleviate the bias that the generated points can not be guaranteed to be a proper subset of the input point cloud. Qian $et\;al.$ \cite{qian2020mops} proposed a matrix optimization-driven network, MOPS-Net, to downsample the point cloud. Different above two works, MOPS-Net leverages the shared MLP to supplement the lack of point-wise local information. Lin $et\;al.$ \cite{lin2021net} combine the K-nearest neighboring algorithm with Local Adjustment (LA), allowing the sampled points have noise immunity characteristics. However, these methods do not guarantee that the learning capacity of sampling network architecture can extract enough features containing context information. Recently, Wang $et\;al.$ \cite{wang2021pst} pioneered a Transformer-based downsampling network PST-NET combined with local-global context information. Although PST-NET acquired competitive performance, the complex structure reduces the expectation of resource saving. In this paper, we design a light-weight Transformer network named LighTN, with favorable FLOPs and Parameters budgets. We will explore this topic in the following chapters.
\section{Method}
\label{sec: method}
In point cloud downsampling scenarios, our goal is to keeps downstream networks can robustly handle specific task based on sampled point sets. To achieve this goal, this paper propose a new framework, LighTN, an end-to-end and task-oriented network. In this section, We first formulate the problem settings and then propose our approach based on a light-weight Transformer architecture. The overall framework of LighTN is present in Figure \ref{fig2}.

\subsection{Problem Settings}
Given an original point cloud $P=\{ {p_i} \in {R^{3 + f}},i = 1,2, \cdots ,N\}$, where $N$ is the number of points and $f$ represents features except three-dimension Cartesian coordinates, the target is get downsampled point cloud $Q$ with $M$ points ($M<<N$) which is an optimal subset distribution of $P$ containing rich context information. The mathematical expression of the above objective function can be expressed in terms of a equation:
 \begin{equation}\label{eq:1}
 {argmin {L_{task}}(F(Q)),M <  < N,M \subset N}
 \end{equation}
where $F( \cdot )$ is the downstream task network and ${L_{task}}$ indicates the task loss. Under this definition, the evaluating indicator of the presented LighTN module becomes explicit, namely minimal performance loss. Unfortunately, although many deep sampling methods have been proposed, three critical problems still hinder the performance of point cloud downsampling simultaneously. \textit{1) differentiable sampling}: during the training and testing phase, learn-based sampling operations need to be end-to-end; \textit{2) learnable capacity}: increase learnable capacity of neural network to encode more critical information with limited storage space; and \textit{3) energy-efficient}: the resource overhead (computation load) of downsampling model must be less than the resource saved by the sampled point cloud through task network.

\textbf{Differentiable sampling}
\label{sec: differentiable sampling}
The downsampling method based on deep learning has demonstrated high accuracy in learning critical points than traditional approaches. However, critical point set generated by downsampling network cannot be guaranteed to be a proper subset in original input. Usually, additional matching operation, such as nearest neighbor search, is required to map each generated point to its nearest neighbor in point cloud input at testing stage. Nevertheless, the matching step restricts further performance improvement in critical point extraction since this additional operation is non-differentiable. Ideally, the critical points generated by neural networks are the proper subset of point cloud $P$. Therefore, this purpose can be denoted as:
\begin{equation}\label{eq:5}
{Q \buildrel \Delta \over = \{ M \in N|dist({q_i},{p_j}) \le \varepsilon ,\lim \varepsilon  \to 0\}}
\end{equation}
where $dist({q_i},{p_j}) \buildrel \Delta \over = \mathop {\sup }\limits_{{q_i} \in M,{p_j} \in N} ||{q_i} - {p_j}|{|_{{R^{3 + f}}}}$ for any subset $q_i$ in $P$. Fortunately, inspired by soft project on weight coordinates \cite{lang2020samplenet}, differentiable relaxation to the matching phase can solve above problem. Based on differentiable relaxation, we introduce a new metric ${L_{soft}}$ that encourages the LighTN to produce an optimal proper subset from the original point cloud. The details of ${L_{soft}}$ will be shown later.

\textbf{Learnable capacity} Point clouds are unordered and irregular, it is challenging to design deep frameworks for learning the relationships between points with permutation invariant characteristics. Intuitively, deeper and wider neural network framework can deposit more information with sufficient learnable parameters to improve the discrimination ability of the model. However, large networks are vulnerable to overfitting when the dataset is insufficient. Besides, The experiments of Zhao $et\;al.$ \cite{zhao2021battle} show that the learning capacity of each network structure is different, such as MLP, CNN and Transformer. This paper will explore the impact of several network structures on point cloud downsampling. See ablation experiments for details.

\textbf{Energy-efficient} Traditional Transformer-based networks sequentially stack multiple Transformer blocks to improve network capacity. Multi-head self-attention is the core component for the Transformer-based block, making it capable to extract ample global context information. Multi-head attention consists of multiple single-head attentions running in parallel. For single-head Scaled Dot-Product self-attention, the computation of point cloud $P$ can be formulated as:
\begin{equation}\label{eq:3}
{SA(P) = F{C_{out}}(Atten(F{C_Q}(P),F{C_K}(P),F{C_V}(P))}
\end{equation}
\begin{equation}\label{eq:4}
{\begin{array}{l}
Atten(Q,K,V) = softmax(\frac{{Q \cdot {K^T}}}{{\sqrt {D/a} }}) \cdot V\\
Q,K \in {R^{N \times (D/a)}},V \in {R^{N \times D}}
\end{array}}
\end{equation}
where $FC( \cdot )$ represents the linear transformation through projection matric, $softmax(\cdot)$ is the activation function, scale ${Q \cdot {K^T}}$ by $1/\sqrt {D/a}$ to improve network stability and $D$ is the dimension of $Q$ and $K$ vectors. Note that $a$ is the scaled factor used to reduce the dimension $D$. The computation cost of single $SA$ is $o(4N{D^2} + 2{N^2}D)$, where $a$ is not considered here. For multi-head self-attention with $m$ heads, the computation cost will be $m \cdot o(4N{D^2} + 2{N^2}D)$. Besides, Transformer always consumes more computation and storage than Convolutional and Full Connection layer of similar structure, since wider network structure and extra components are introduced, i.e., positional encoding and feed forward networks. Therefore, the framework of Transformer needs to be optimized under limited resource constraints.

Ideally, a task network introducing an independent downsampling network with reasonable downsampling ratio will consume less resource overhead than a single task network running on original point cloud. In theory, we divide the resource overhead into computation load $C$ and storage space $S$, respectively:
\begin{equation}\label{eq:2}
{B(({C_{{L_{N \to M}}}} + {C_{{T_M}}}),({S_{{L_{N \to M}}}} + {S_{{T_M}}})) < B({C_{{T_N}}},{S_{{T_N}}})}
\end{equation}
where $C_{{T_N}}$ and $C_{{T_M}}$ denotes computation load of task network on $N$ and $M$ point sets respectively, $C_{{L_{N \to M}}}$ means the computation cost of LighTN with $N$ input points that output $M$ critical points. Similarly, the S has consistent representation as $C$. $B( \cdot , \cdot )$ is a pre-defined resource limit metric.
\subsection{Differentiable Soft Projection}
As discussed in \ref{sec: differentiable sampling}, it is necessary to eliminate the point cloud matching algorithm in test phase for end-to-end learning. In this paper, we utilize the soft projection operation \cite{lang2020samplenet}, denoting the average weight of the $k$ nearest neighbors points of ${q_i}$ as soft projected point $z$ to represent ${q_i}$. Hence, soft projected point $z$ can be denoted as:
\begin{equation}\label{eq:6}
{z = \sum\limits_{{p_i} \in {{\cal N}_k}({q_i})} {{w_i} \cdot {p_i}}}
\end{equation}
Ulteriorly, Gumbel-Softmax Trick is used to optimize the constraints in Eq.\ref{eq:5}:
\begin{equation}\label{eq:7}
{{w_i} = \frac{{{e^{ - dist_i^2/t}}}}{{\sum\limits_{j = 1}^k {{e^{ - dist_j^2/t}}} }}}
\end{equation}
where $t$ is a learnable temperature coefficient that controls the distribution shape of the weight $w_i$. It is clear that when $t \to {0^ + }$, point $z$ is approximately considered to be the proper subset of input point cloud. In our sampling method, we add a project loss in sampling loss:
\begin{equation}\label{eq:8}
{{L_{soft}} = T(t),t \in [0, + \infty )}
\end{equation}
where $T( \cdot )$ as a function of $t$ that control its nonlinearities of the projection loss. The experimental results demonstrate the exponential function $exp( \cdot )$ guides the neural network to converge to higher performance. The details are presented in ablation study.

\subsection{Light-weight Transformer}
\label{sec: Light-weight Transformer}
Recently, Transformer's learning capacity has been proved to effectively capture more useful relationship features between points in terms of point cloud shape and geometric dependencies. A standard Transformer framework contains six major components, \textit{1)} positional encoding; \textit{2)} input embedding layer; \textit{3)} multi-head self-attention block; \textit{4)} Feed Forward Networks; \textit{5)} layer normalization; and \textit{6)} skip connection. To overcome the limitations of the resource constraints, we design a light-weight Transformer network with a self-correlation mechanism. The lower left corner of Figure \ref{fig2} shows the overall framework of LighTN.

\textbf{Vanishing position embedding} In 2D image recognition, position encoding is an essential mechanism for preserving the local relative position of patches, conducive to improving network performance \cite{han2021transformer,dosovitskiy2020image}. However, the arrangement of point clouds has no fixed order under the irregular and unordered characteristics. Meantime, 3D coordinate is a substitute for position encoding because it includes natural spatial location information. Therefore, to reduce the storage and computing overhead of LighTN, we remove the position encoding block directly. As a result, this removal operation eliminates the overhead of positional encoding itself, and does not increase the feature dimension in the propagation process.

\textbf{Light-weight input embedding block} Given a sequence of $N$ points with $3+f$ dimensional features, these inputs are first fed to the input embedding layer to learn a ${d_o}$-dimensional embedded features ${F_O} \in {R^{N \times {d_o}}}$. The purpose of input embedding layer is to embed the original point cloud into a high-dimensional feature space which facilitate subsequent learning. In this paper, we use a shared linear layer as input embedding layer and empirically set ${d_o} = 64$. Compared to the computationally-saving input embedding setting in \cite{guo2021pct}, our design has fewer layers and halved dimension. The computational costs for the input embedding in the Transformer block \cite{guo2021pct} and the LighTN block are ${\cal O}(2Nd_m^2)$ and ${\cal O}(Nd_o^2)$ respectively, where $ {{d_o} = \frac{1}{2}{d_m}}$.
\begin{table*}[t]
    \centering
    \caption{Classification accuracy with different downsampling methods.}
    \renewcommand{\arraystretch}{1.15}
    \begin{tabular}{c|c|c|c|c|c|c|c|c|c}
        \hline
        m & Voxel \cite{qian2020mops} & RS \cite{dovrat2019learning} & FPS \cite{dovrat2019learning}& S-NET \cite{dovrat2019learning}& PST-NET \cite{wang2021pst}& SampleNet \cite{lang2020samplenet}& MOPS-Net \cite{qian2020mops}& DA-Net \cite{lin2021net} & LighTN (Ours) \\ \hline
        512 &73.82 &87.52 &88.34 &87.80 &87.94 &88.16 &86.67 &89.01 &\textbf{89.91}  \\
        256 &73.50 &77.09 &83.64 &82.38 &83.15 &84.27 &86.63 &86.24 &\textbf{88.21}  \\
        128 &68.15 &56.44 &70.34 &77.53 &80.11 &80.75 &86.06 &85.67 &\textbf{86.26}  \\
        64  &58.31 &31.69 &46.42 &70.45 &76.06 &79.86 &85.25 &85.55 &\textbf{86.51}  \\
        32  &20.02 &16.35 &26.58 &60.70 &63.92 &77.31 &84.28 &85.11 &\textbf{86.18}  \\
        16  &13.94 &7.15  &13.29 &36.16 &42.29 &51.09 &\textbf{81.40} &-     &79.34  \\
        8   &3.85  &3.27  &3.47  &20.81 &19.32 &23.94 &52.39 &-     &\textbf{52.92}  \\
        4   &-     &1.43  &1.63  &5.47  &5.40  &5.55  &-     &-     &\textbf{22.08}  \\
        2   &-     &1.22  &1.02  &2.80  &3.57  &1.45  &-     &-     &\textbf{7.78}   \\ \hline
    \end{tabular}
\label{table1}
\end{table*}

\textbf{Single head self-correlation layer}
Most works introduce multi-head attention mechanism to improve the learning capacity of Transformer networks. However, this mechanism is resource-intensive that seriously affects the energy efficiency of the point cloud downsampling task, as shown in Eq.\ref{eq:2}. Especially when the energy consumption of the task network itself is relatively small, the expansion of downsampling network structure may lead to the improvement of the overall overhead. To overcome this problem, we propose a self-correlation block to model the contextual relationships between these $N$ input point sets with ${d_o}$ features, as shown in the lower right corner of Figure \ref{fig2}. We think downsampled point cloud has the property of natural fault tolerance because there are many approximate solutions with the increase of point distribution density. Here, we hypothesize the learning ability of a single self-attention layer allows LighTN to extract enough relationships between points with permutation invariant characteristics. Experimental results in ablation studies support this view. Furthermore, the computation overhead of dot-product attention block is controlled by two components: $1)$ the projection matrices of query $Q$, key $K$ and value $V$; and $2)$ the matrix dot-product operation, as shown in Eq.\ref{eq:3} and Eq.\ref{eq:4}. Recently, the work of Mobile-Former \cite{Chen2021MobileFormerBM} removed the projection matrices of both ${W^K}$ and ${W^V}$, or ${W^Q}$ to save computations. Inspired by this work, we eliminate the ${W^K}$, ${W^V}$ and ${W^Q}$ simultaneously to further improvement over the computation and storage consumptions. Due to the process of calculating the attention score only involving input self-correlation parameter operations, we name it self-correlation layer. Specifically,
\begin{figure}[!hbt]
    \centering
    \includegraphics[scale=0.7]{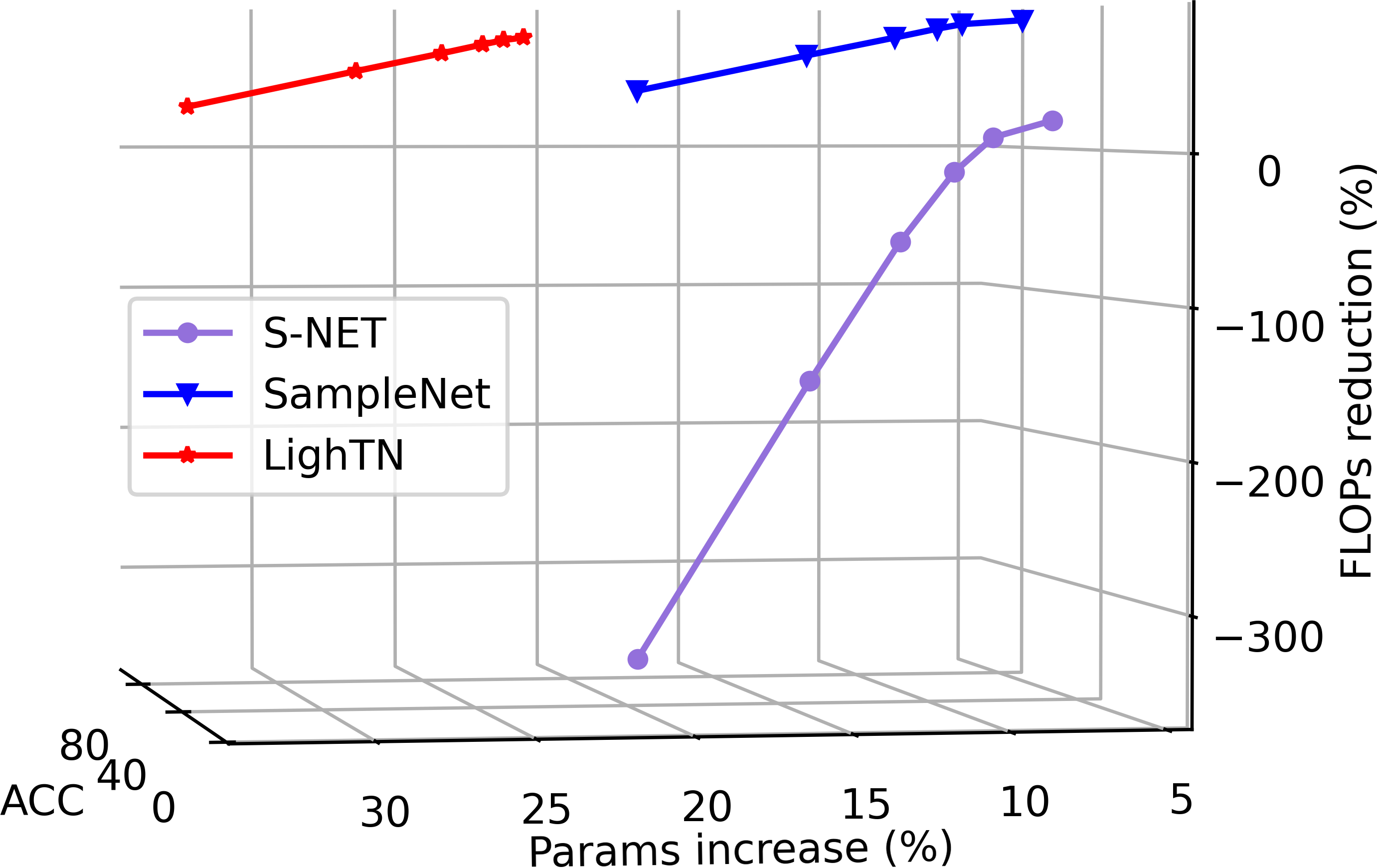}
    \caption{Computation (FLOPs) reduction, parameter (Params) increase and accuracy (ACC) comparison. The 6 mark points from left to right in each line correspond to downsampling ratios \{2, 4, 8, 16, 32, 64\} respectively. All data are obtained by testing the whole model, including downsampling and task networks.}
    \label{fig3}
\end{figure}
Eq.\ref{eq:3} and Eq.\ref{eq:4} are modified as follows:
\begin{equation}\label{eq:9}
{SA(X) = F{C_{out}}(C(X))}
\end{equation}
\begin{equation}\label{eq:10}
{C(X) = softmax(\frac{{X \cdot {X^T}}}{{\sqrt {D} }}) \cdot X}
\end{equation}
where $X$ is the output of input embedding block. Noticeably, the output of $X \cdot {X^T}$ is a symmetric matrix $A$ that satisfies ${A^T} = A$, where ${A^T}$ denotes the transpose. Unlike natural language processing tasks in which the order between two words is one of the key characteristics that express different meanings, the sequence between two points are interchangeable under the disorder characteristic. Definitely, the attention scores between two points in the symmetric matrix satisfies the permutation invariant characteristic that the elements $a$ of $A$ have the form ${a_{ij}} = {a_{ji}}$, where $i$ and $j$ denote two different points. The computational cost for self-correlation is only ${\cal O}(Nd_o^2 + 2{N^2}{d_o})$.

\textbf{FFN scaling} Replacing the multi-head attention with single-head self-correlation layer greatly reduces the resource overhead of the Transformer network, but the learnable parameters are also decreased. To solve this problem, in the work of Mehta $et\;al.$\cite{DBLP:conf/iclr/MehtaGIZH21}, a group linear transformations (GLTs) is inserted in front of the attention module to increase the network depth. Different from their work, we introduce expand-reduce strategy\cite{DBLP:conf/iclr/MehtaGIZH21} in FFN. Compared with a standard FFN block with two linear layers, our method simply and efficiently increases the depth and learnable parameters of LighTN with small resource overhead. Through experiments, We build three linear layers and use expand-reduce strategy in middle layer, which reduces the dimensionality of the input from ${d_f}$ to ${d_f}/r$ while guaranteeing performance. In this paper, we set the reduction ratio $r=2$. Thus, the FFN only increases ${\cal O}(Nd_f^2/4)$ computational cost.

\subsection{Final Loss Function}
\label{sec: Final Loss Function}
To learn the simplified point sets by LighTN, we present a novel loss function, as shown in Figure \ref{fig2}, which consists of two components ${L_{sampling}}$ and ${L_{task}}$. We advocate that LighTN can learn simplified point sets that are the proper subset of original point cloud. To enforce this, we introduce Chamfer Distance (CD) function. CD distance between input point $P$ and simplified point $Q$ is defined as:
\begin{equation}\label{eq:11}
{{L_{CD}}(Q,P) = \frac{1}{Q}\sum\limits_{q \in Q} {\mathop {\min }\limits_{p \in P} } ||q - p||_2^2 + \frac{1}{P}\sum\limits_{p \in P} {\mathop {\min }\limits_{q \in Q} } ||p - q||_2^2}
\end{equation}
${L_{CD}}$ can promote $Q$ to become the nearest point in $P$. One of the main limitations of ${L_{CD}}$ is that they are oblivious to uniform distribution of simplified point sets, making it challenging for global surface representation. To alleviate above problem, we adopt a repulsion loss \cite{yu2018ec} which encourages more uniform distribution of the simplified points on the original point cloud. Thus, we define the repulsion loss as follows:
\begin{equation}\label{eq:12}
{{L_{repl}}(Q) = \frac{1}{{Q \cdot k}}\sum\limits_{1 \le {q_i} \le Q} {\sum\limits_{{{q'}_i} \in {{\cal N}_k}({q_i})} {\eta (||{{q'}_i} - {q_i}|{|_2})} } }
\end{equation}
where $\eta (r) = \max (0,{h^2} - {r^2})$ is a function to guarantee the $q$ is not too close to others in $Q$, where $h$ is empirically set as 0.001 and $k$ is set as 15. According to the discussion above, the total sampling loss is:
\begin{equation}\label{eq:13}
{{L_{sampling}} = {L_{CD}} + \alpha {L_{repl}} + \beta {L_{soft}}}
\end{equation}
where $\alpha$ and $\beta$ are regularization parameters.

${L_{task}}$ is a given loss function corresponding to a priori defined network with task-specific requirements. Specifically, we do not change the ${L_{task}}$ during LighTN training. Therefore, putting all loss functions together, we propose to minimize:
\begin{equation}\label{eq:14}
{argmin{L_{sampling}}(P,Q) + \delta {L_{task}}(Q)}
\end{equation}
where $\delta$ is balancing weight.
\begin{figure*}[!hbt]
    \centering
    \includegraphics[scale=0.6]{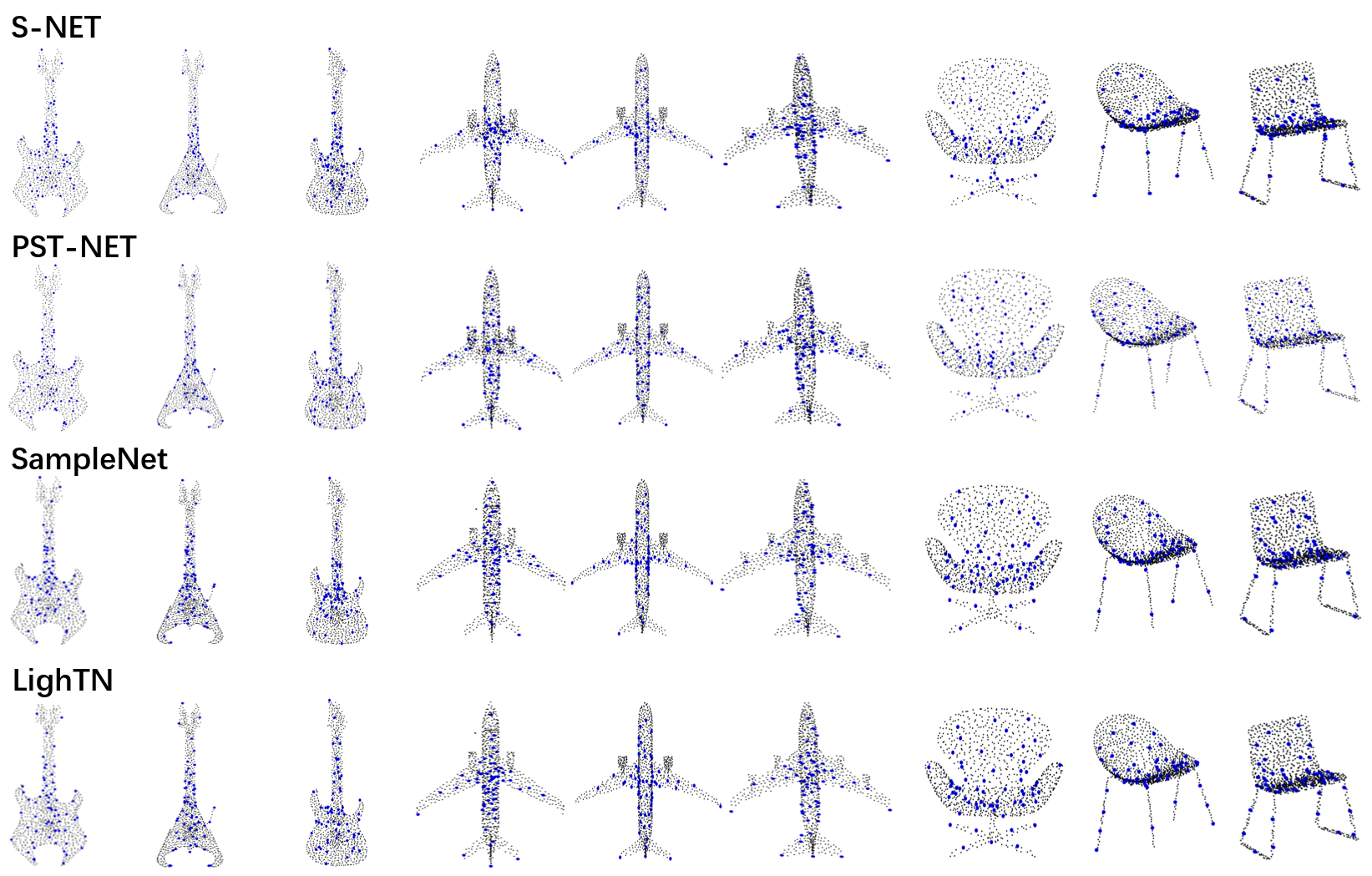}
    \caption{Visualization of the sampled points of classification task learned by different downsampling methods on ModelNet40 dataset. Original point cloud contains 1024 points (in gray), downsampling ratio is set to $m = 16$. The sampled 64 points mark with enlarged blue dots. Three classes of objects are shown: Guitar, airplane and chair.}
    \label{fig4}
\end{figure*}
\section{Experiments}
\label{sec: experiments}
This section evaluates the performance and resource overhead of LighTN for the task-oriented point cloud downsampling. Ideally, LighTN is a plug-and-play module that can be combined with any point cloud processing framework that requires downsampling. In this research, We explore the model performance of LighTN on two  machine learning tasks: classification and registration, respectively. In our experiments, the proposed LighTN is compared with a series of state-of-the-art downsampling methods, including \textit{1) commonly used traditional methods:} FPS, random sampling and voxel; and \textit{2) task-oriented methods:} simplified methods with non-differentiable matching operation (S-NET \cite{dovrat2019learning} and PST-NET \cite{wang2021pst}); simplified methods with differentiable relaxation matching operation (SampleNet \cite{lang2020samplenet} (commonly used baseline), DA-Net \cite{lin2021net} and MOPS-Net \cite{qian2020mops}).

\subsection{Dataset and Metric}
The classification and registration tasks are tested on ModelNet40 \cite{wu20153d} dataset. ModelNet40 contains 12311 manufactured 3D CAD models in 40 common object categories, i.e., airplane, bed and door. For a fair comparison, we leverage the official train-test split strategy with 9840 CAD models for training stage and 2648 CAD models for testing stage. In order to obtain the 3D Cartesian coordinates of each CAD model, uniformly sample approach \cite{qi2017pointnet} is used to extract 1024 points on mesh faces. More specifically, we use the XYZ-coordinate as point cloud input without other attributes. For evaluation metrics, the performance of LighTN is the accuracy and rotation error for classification and registration tasks, respectively. Computation load and storage space are used to represent the resource overhead. Besides, the downsampling ratio is defined as $N/m$, where $N$ is the number of original points and $m$ represents the number of downsampled points.

\subsection{Experiments on Classification}
\textbf{Implementation details} Following the experiments of S-NET \cite{dovrat2019learning}, PST-NET \cite{wang2021pst}, SampleNet \cite{lang2020samplenet} and MOPS-Net \cite{qian2020mops}, we use the PointNet \cite{qi2017pointnet} as our task network to perform point cloud classification. We implement the LighTN with PointNet in Tensorflow \cite{abadi2016tensorflow}. The GPU version is Tesla V100 with 16G memory. For LighTN, we use the Adam optimizer with mini-batch size of 32 and initial learning rate of 0.01. In order to ensure the performance of PointNet is not disturbed in training and test phases, we adopt the original network configuration and hyperparameters without any changes made.

\begin{table*}[t]
    \centering
    \caption{Rotation error with different downsampling methods.}
    \renewcommand{\arraystretch}{1.15}
    \begin{tabular}{c|c|c|c|c|c|c|c}
        \hline
        m & Voxel \cite{qian2020mops} & RS  & FPS &  SampleNet \cite{lang2020samplenet}& MOPS-Net \cite{qian2020mops}& LighTN (non-FPS) &LighTN (Ours) \\ \hline
        512 &-     &5.04  &4.64  &4.22  &-     &4.60  &\textbf{4.11}  \\
        256 &-     &4.85  &4.42  &4.53  &-     &4.69  &\textbf{4.32}  \\
        128 &-     &6.54  &4.82  &4.85  &-     &4.92  &\textbf{4.60}  \\
        64  &10.37 &17.32 &5.65  &5.44  &8.58  &5.27  &\textbf{5.23}  \\
        32  &14.46 &17.90 &7.29  &5.68  &7.97  &5.86  &\textbf{5.60}  \\
        16  &-     &22.02 &9.84  &6.90  &-     &\textbf{6.28} &6.36  \\
        8   &44.80 &30.25 &14.37 &10.09 &12.64 &\textbf{8.09} &8.83  \\ \hline
    \end{tabular}
\label{table2}
\end{table*}

\textbf{Performance and efficiency}
\begin{figure}[!hbt]
    \centering
    \includegraphics[scale=0.7]{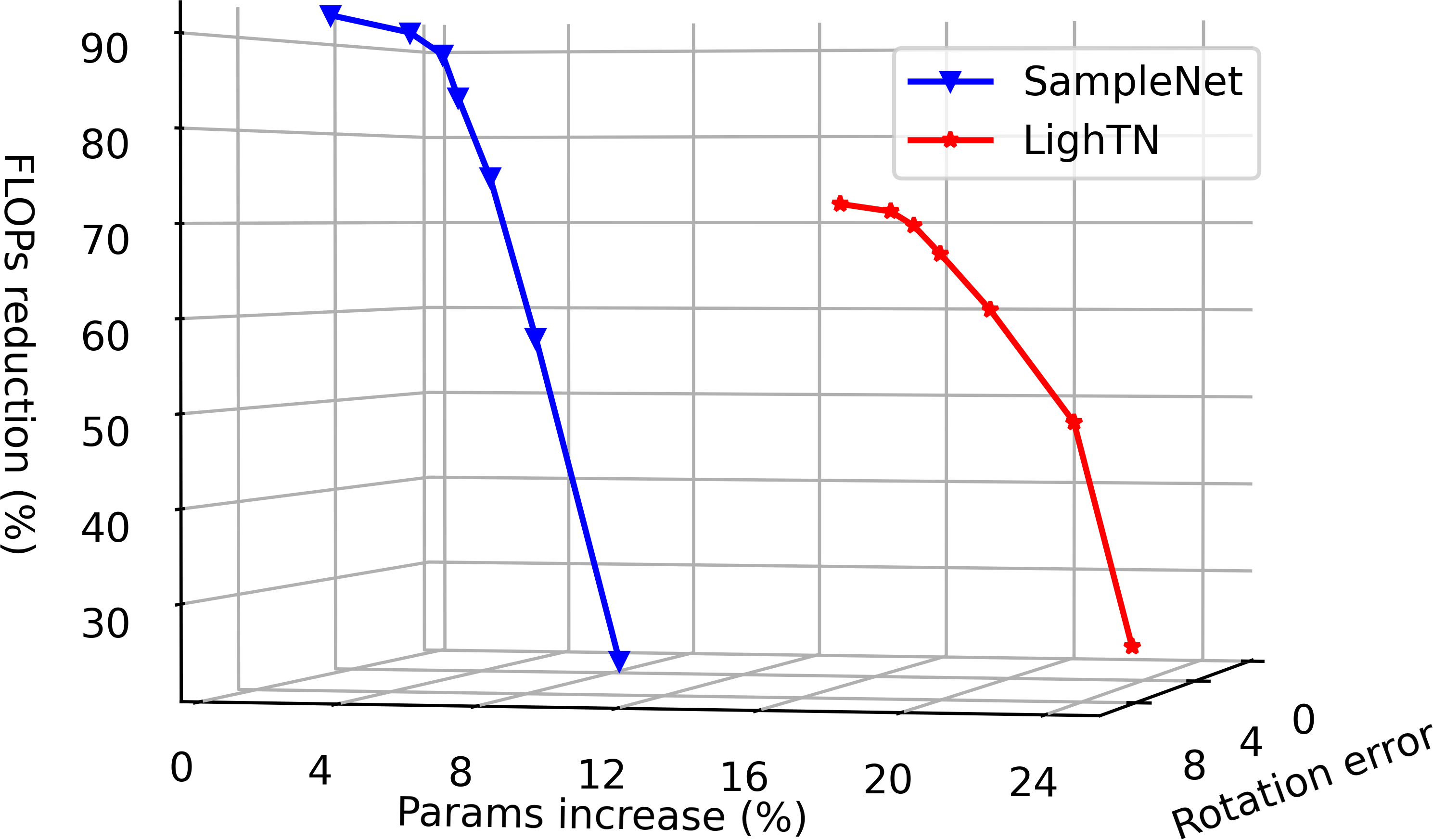}
    \caption{Computation (FLOPs) reduction, parameter (Params) increase and rotation error. The 6 mark points from right to left in Params line correspond to downsampling ratios \{2, 4, 8, 16, 32, 64\} respectively. All data are obtained by testing the whole model, including downsampling and task networks.}
    \label{fig5}
\end{figure}
The classification results on ModelNet40 are reported in Table \ref{table1}. Firstly, nearly all experiments show that the output accuracy of task network do not change significantly when the number of point set is sampled from 1024 to 512. This result demonstrates the number of points entered into the task network, PointNet, is redundant. Therefore, invariant output accuracy with lower overhead costs can be achieved by using downsampling methods on input point cloud for a specific task network. Secondly, the output accuracy decreases rapidly with the increase of downsampling rate for task-irrelevant methods. Besides, the performance obtained by these two-stage methods is often suboptimal because of the algorithm's randomness and the reproducibility issue. Thirdly, task-oriented methods achieve higher classification accuracy around all downsampling rates than traditional point cloud simplified approaches. S-NET and PST-NET obtain competitive performance. For example, after the input points down to 256 (reduce points by 75\%), the output accuracy of the PointNet remains above 82\%. However, as the number of points declines, the accuracy drops rapidly. The reason might be that the additional matching process from simplified points to input point cloud introduce deviation, especially when the number of sampled points itself is small, slight deviation will cause large performance drop. Fourthly, we see that both SampleNet, MOPS-Net, DA-Net achieve good performance, and our LighTN achieves superior accuracy. This result further proves that the proposed task-oriented end-to-end mechanism has an excellent generalization ability, and it is more suitable for point cloud downsampling tasks. On the other hand, both S-NET, SampleNet and MOPS-Net use PointNet as basic point cloud simplified network. Limited by the learning ability of PointNet, their performance is not the best. This indicates that our simplified network framework has a higher feature learning ability to boost the accuracy further as the number of points continues to decline.

Figure \ref{fig3} presents the parameters increase versus computation reduction. We introduce floating-point operation (FLOPs) metric to measure the computation load. Memory space is used to represent learnable parameters. Significantly, reported resource overhead data is for the whole execution process, which includes the point cloud input going through the LighTN and PointNet in turn to get the classification result. Evaluations show that LighTN achieves competitive results. For example, set the downsampling rate to 32 (Only 32 points simplified by LighTN passes through the PointNet), the whole execution process has 223.2M FLOPs and 4.24M parameters. Compared with the PointNet with 1024 input points (the original cost is 927.2M FLOPs and 3.48M parameters), above setting reduces 75.93\% FLOPs with only 21.91\% parameters increase, and the output accuracy of PointNet remains at 86.18\% (accuracy reduction of approximately 4\% on average). Note that in S-NET the operation of whole execution process consumes more FLOPs than original task network with the input of 1024 points, which is seriously inconsistent with the original intention of downsampling.

\textbf{Visualization results}
We visualize the simplified results of point clouds for 1024 input points with downsampling ratio of 16 over three categories: guitar, airplane and chair. As shown in Figure \ref{fig4}, non-differential-based S-NET pays more attention to the main structure of the object and does not cover the prominent regions well. In contrast, PST-NET and SampleNet achieve better downsampling, but the visualize images show that some prominent regions have insufficient or missing attention. Meantime, above three approaches expose the problem that point cloud distribution is relatively concentrated which cannot cover the entire surface of the object well. For this reason, LighTN utilizes the repulsion loss to encourage more uniform distribution of the simplified points. The visualization results of the last row demonstrate the proposed LignTN has the best downsampling ability to simplify the point clouds with better task-oriented output accuracy.
\begin{figure*}[!hbt]
    \centering
    \includegraphics[scale=0.5]{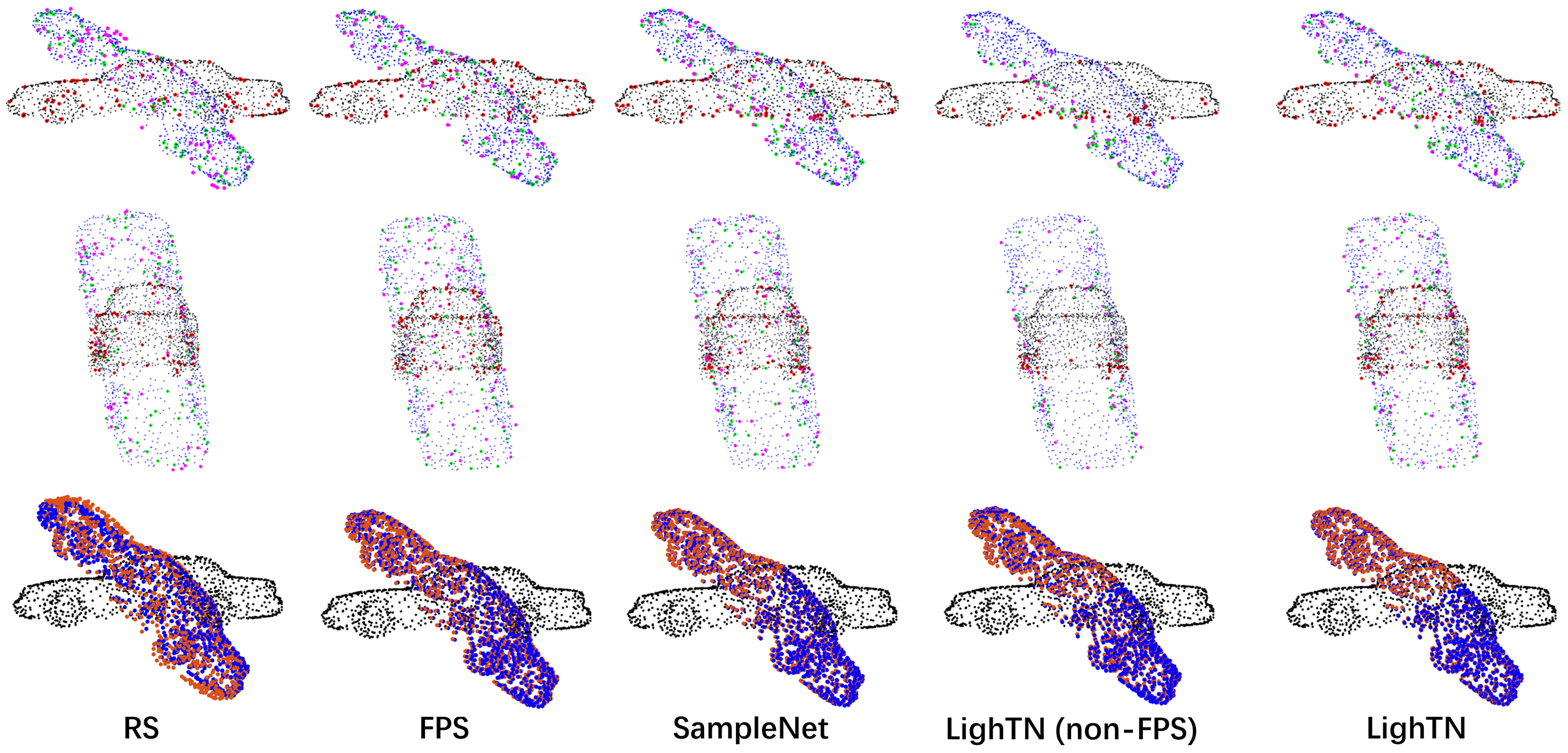}
    \caption{Visualization of the sampled points of the registration task with $m=16$. \textbf{Upper}: template (in blue) and source (in black) point clouds from the side view. Sampled 64 points of a template and a source point cloud are marked with enlarged green and red dots, respectively, and then the transformed sampled source points are marked with magenta. \textbf{Middle}: registration sampled results from the top view of template point cloud. \textbf{Bottom}: the alignment between source point cloud with 1024 points and ground truth. The transformation matrix between two original point clouds is obtained by sampled predicted.}
    \label{fig6}
\end{figure*}
\subsection{Experiments on Registration}
\textbf{Implementation details}
Following the experiments of SampleNet and MOPS-Net, we adopt the PCRNet \cite{sarode2019pcrnet} as task network of point cloud registration for fair comparisons. We implement the LighTN and PCRNet in PyTorch \cite{paszke2019pytorch}. Unlike the hyper-parameter setting in classification task, we set the initial learning rate to 0.001 for LighTN. On the other hand, we still use the official hyper-parameter setting during the training stage for PCRNet. In this section, we test the performance of Car category (each car consists of 1024 points) in ModelNet40 dataset.
\begin{figure}[!hbt]
    \centering
    \includegraphics[scale=0.7]{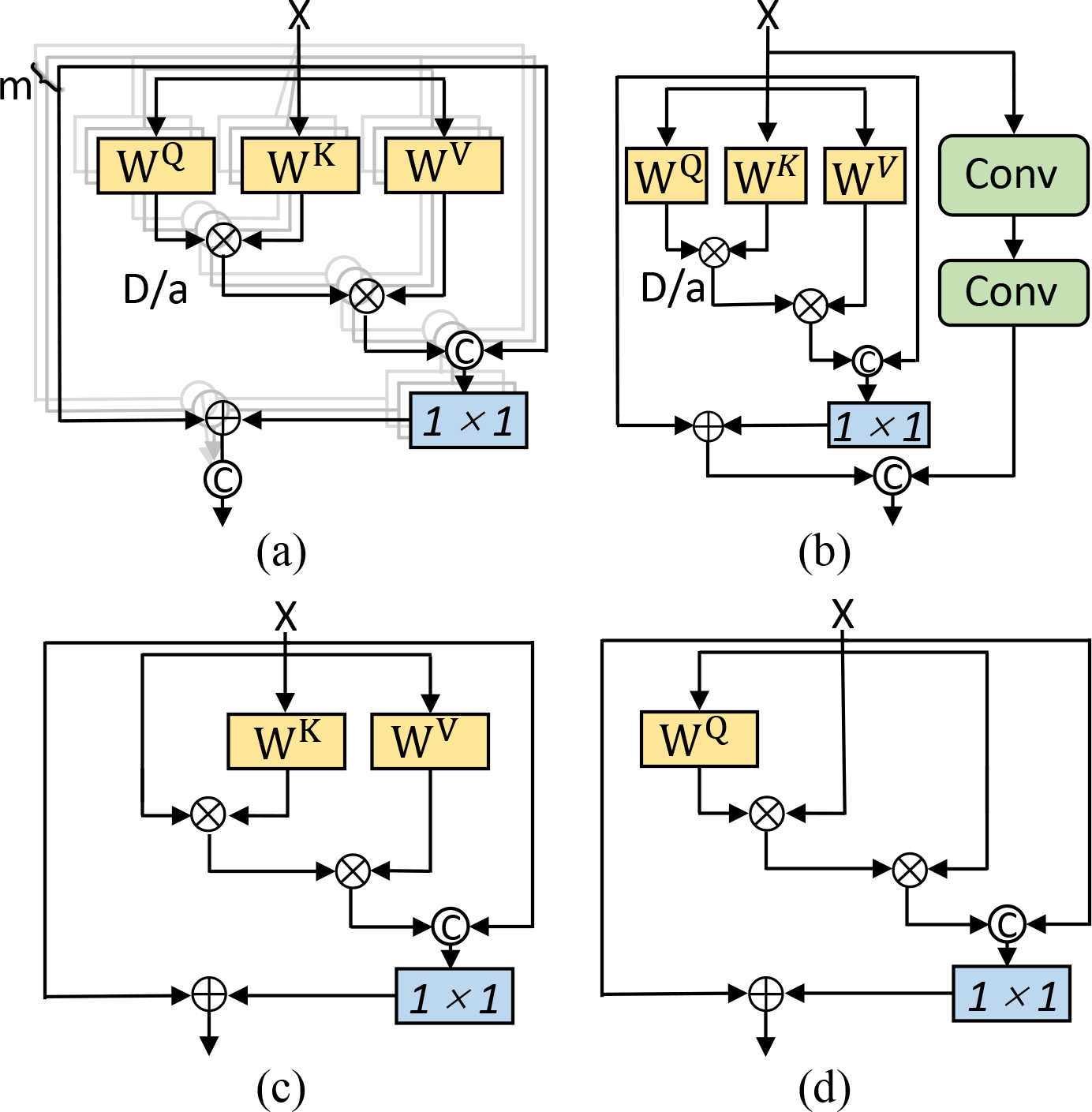}
    \caption{Ablation experiments for different self-attention blocks. (a) Multi-head self-attention with projection matrices ${W^Q}$, ${W^K}$ and ${W^V}$ where $m$ is the number of heads, $D$ is the dimension of $Q$ and $K$ vector and $a$ is the scaled factor, as described in Eq.\ref{eq:4}; (b) Single-head self-attention block with local convolution layer (Conv); (c) single-head self-attention block where ${W^Q}$ is removed to save computations; (d) single-head self-attention block where ${W^K}$ and ${W^V}$ are removed.}
    \label{fig7}
\end{figure}
\begin{figure}[!hbt]
    \centering
    \includegraphics[scale=0.7]{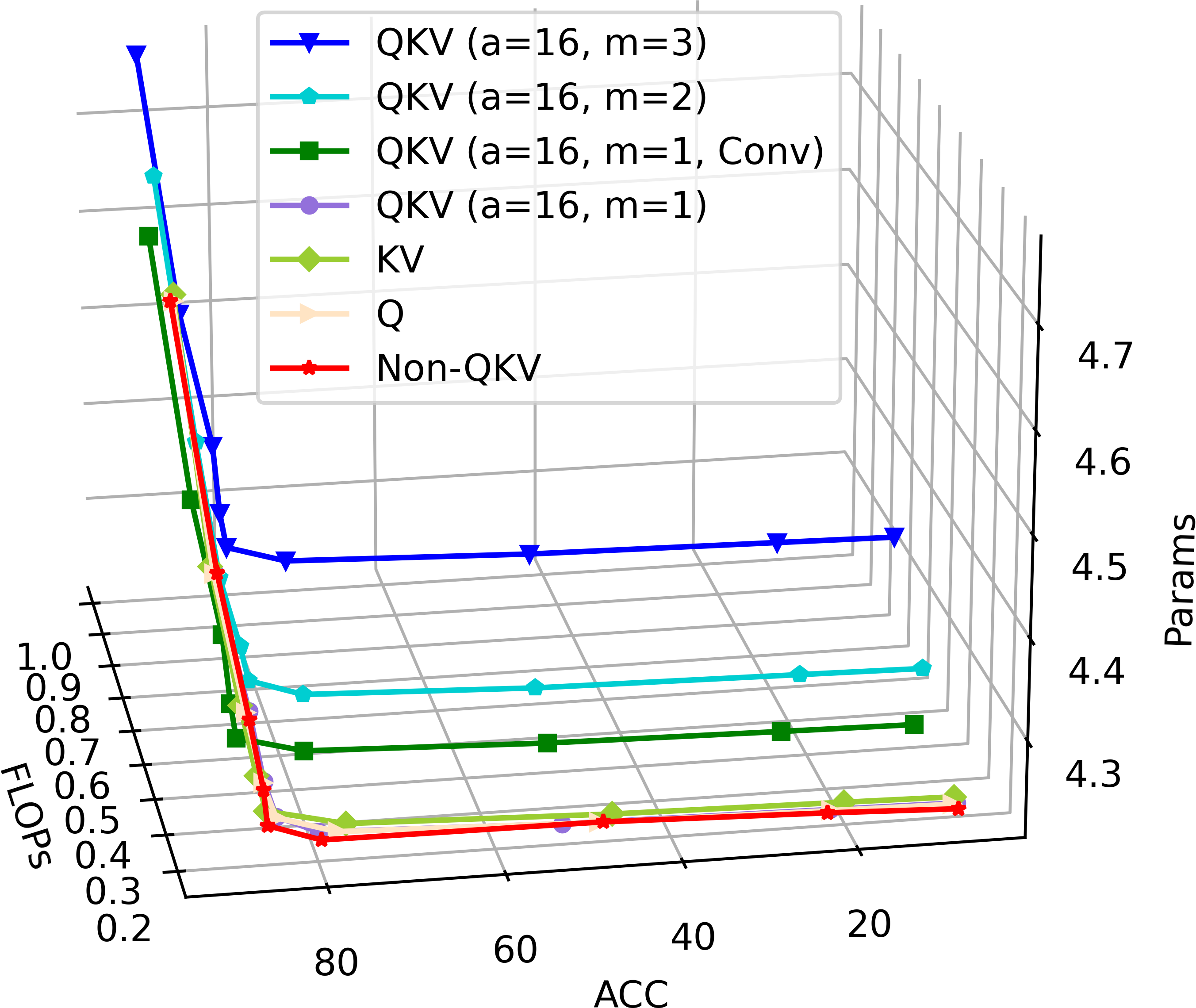}
    \caption{Comparison of self-attention blocks on ModelNet40 classification. The $QKV$ indicates the attention mechanism with complete projection matrices. $a$ is the scaled factor, $m$ represents the number of heads and $Conv$ is convolutional operation. $KV$ and $Q$ represent modules of Figure \ref{fig7}(c) and \ref{fig7}(d), respectively. $Non-QKV$ is our designed self-correlation block. Note that the 9 mark points from left to right of each line on ACC axis correspond to downsampling ratios \{2, 4, 8, 16, 32, 64, 128, 256, 512\} respectively.}
    \label{fig8}
\end{figure}

\textbf{Performance and efficiency}
Point cloud registration is the mapping process of finding a spatial transformation that aligns two point clouds in 3D space. In this paper, the evaluation metric used is mean rotation error (MRE), which assesses how imprecisely the predicted rotation is aligned to ground truth. Table \ref{table2} shows the registration results on Car category. The results show that as the downsampled points are reduced to 64 or lower, traditional task-irrelevant methods suffer severe performance degradation because fewer points are difficult to represent the global characteristics of the original point cloud. In contrast, LighTN achieves the best alignment results in all tests, which shows better learning ability of task-oriented downsampling. Besides, to ensure the sampling operation will not sample the same points in the test stage, we remove duplicate points and leverage FPS to complete the number of points. Comparison results with and without FPS (non-FPS) show that FPS-based completion operation can further improve the network performance. Specifically, if there is no clear explanation in this paper, FPS-based completion strategy is adopted in all experiments.
\begin{figure}[!hbt]
    \centering
    \includegraphics[scale=0.7]{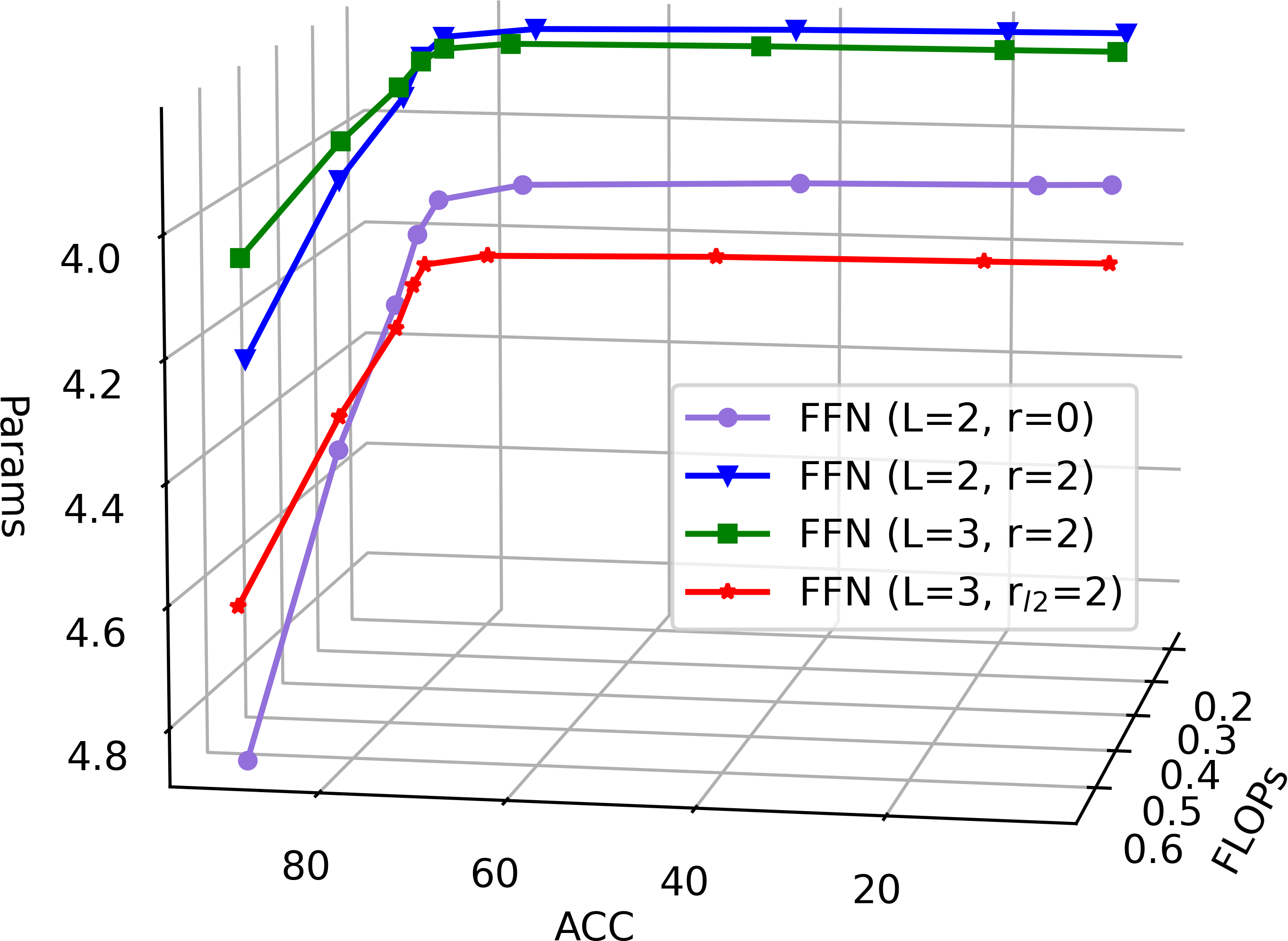}
    \caption{Ablation experiments for different FFN architectures on classification task. $L$ and $r$ denote the number of layers and reduction ratio, respectively. ${r_{l2}}$ means that expand-reduce strategy is only applied in the middle layer.
    The 9 mark points from left to right of each line on ACC axis correspond to downsampling ratios \{2, 4, 8, 16, 32, 64, 128, 256, 512\} respectively.}
    \label{fig9}
\end{figure}

Figure \ref{fig5} shows the parameters increase versus computation reduction. LighTN achieves competitive results compared to SampleNet, which uses the low-overhead PointNet as downsampling network. For example, set the downsampling ratio to 32, the FLOPs reduce 69.73\% with only 17.97\% parameters increase, while the mean rotation error remains at 5.60$^\circ$.

\begin{table}[t]
    \centering
    \caption{Ablation study: a controlled comparison of different loss functions of classification task on ModelNet40.}
    \renewcommand{\arraystretch}{1.15}
    \begin{tabular}{c|c c c }
        \hline
        \tabincell{l}{m}   & \tabincell{c}{${{\rm{L}}_{{\rm{CD}}}}$ \\ + ${{\rm{L}}_{{\rm{soft}}}}({t^2})$ \cite{lang2020samplenet} } & \tabincell{c}{${{\rm{L}}_{{\rm{CD}}}}$ \\ + ${{\rm{L}}_{{\rm{soft}}}}({e^t})$}   &\tabincell{c}{${{\rm{L}}_{{\rm{CD}}}}$\\+${{\rm{L}}_{{\rm{soft}}}}({e^t})+{{\rm{L}}_{{\rm{repl}}}}$} \\ \hline
         512&89.74     &89.14   &\textbf{89.91}  \\
         256&87.88     &87.93   &\textbf{88.21}  \\
         128&86.14     &\textbf{86.62}   &86.26  \\
         64 &85.45     &86.26   &\textbf{86.51}  \\
         32 &85.45     &85.41   &\textbf{86.18}  \\
         16 &\textbf{79.61}     &78.16   &79.34  \\
         8  &47.69     &51.17   &\textbf{52.92}  \\
         4  &22.08     &\textbf{22.89}   &22.08  \\
         16 &7.13      &7.33   &\textbf{7.78}    \\ \hline
    \end{tabular}
\label{table3}
\end{table}
\textbf{Visualization results}
The downsampling ratio is set to 16. Figure \ref{fig6} visualizes the point cloud registration results of LighTN compared with other baselines. Although FPS achieves competitive results, it is not sensitive to prominent regions, such as the lower contour of the wheel. On the other hand, learn-based approaches are better than traditional methods. That shows the advantages of end-to-end and differentiable training. As illustrated, LighTN successfully focuses on both edge information (vehicle contour) and prominent regions, proving that our work is more favorable for performance.

\subsection{Ablation Study}
To probe the validity of the specific designs in LighTN, we conducted several controlled experiments.

\textbf{Self-attention configuration}
To demonstrate the effectiveness of the self-correlation block designed in section \ref{sec: Light-weight Transformer}, we conduct comparative experiments on classification tasks of self-attention blocks proposed by several works in terms of accuracy (ACC), computation load (FLOPs) and storage space (Params). As depicted in Figure \ref{fig7}(a), we choose multi-head dot-production attention block as our baseline approach for the evaluation. It has been pointed out that the framework based on self-attention mechanism improves the global feature extraction ability. Furthermore, wang $et\;al.$ \cite{wang2021pst} introduced a local-global attention mechanism to capture fine-grained local features, which is crucial for vision models. Inspired by their work, we present a local-global single-head attention block with two convolutional layers, as shown in Figure \ref{fig7}(b). Note that the resource overhead of local block is positively correlated with the number of convolutional layers. On the other hand, projection matrices have been proved that it can be removed for saving computations \cite{Chen2021MobileFormerBM}, so we designed two light-weight dot-production attention blocks (see Figure \ref{fig7}(c) and Figure \ref{fig7}(d)).

\begin{table*}[t]
    \centering
    \caption{Ablation study: a controlled comparison of temperature functions $T$ for registration task.}
    \renewcommand{\arraystretch}{1.15}
    \begin{tabular}{c|c| c c c c c}
        \hline
        \tabincell{l}{m}
        &\tabincell{c}{${{\rm{L}}_{{\rm{CD}}}}$ \\ + ${{\rm{L}}_{{\rm{soft}}}}({t^2})$ \cite{lang2020samplenet} }
        &\tabincell{c}{${{\rm{L}}_{{\rm{CD}}}}$\\+${{\rm{L}}_{{\rm{soft}}}}({t})+{{\rm{L}}_{{\rm{repl}}}}$}   &\tabincell{c}{${{\rm{L}}_{{\rm{CD}}}}$\\+${{\rm{L}}_{{\rm{soft}}}}({t^2})+{{\rm{L}}_{{\rm{repl}}}}$}  &\tabincell{c}{${{\rm{L}}_{{\rm{CD}}}}$\\+${{\rm{L}}_{{\rm{soft}}}}({t^3})+{{\rm{L}}_{{\rm{repl}}}}$}  &\tabincell{c}{${{\rm{L}}_{{\rm{CD}}}}$\\+${{\rm{L}}_{{\rm{soft}}}}({t^4})+{{\rm{L}}_{{\rm{repl}}}}$}  &\tabincell{c}{${{\rm{L}}_{{\rm{CD}}}}$\\+${{\rm{L}}_{{\rm{soft}}}}({e^t})+{{\rm{L}}_{{\rm{repl}}}}$} \\ \hline
        512 &4.22 & 4.22  &4.17     &4.16     &4.25     &\textbf{4.14} \\
        256 &4.43 & 4.35  &4.36     &4.42     &4.44     &\textbf{4.33} \\
        128 &4.82 & 4.76  &4.85     &4.79     &4.73     &\textbf{4.65} \\
        64  &5.41 & 5.32  &5.39     &5.43     &\textbf{5.26}     &\textbf{5.26} \\
        32  &5.82 & 5.80  &\textbf{5.75}      &5.89     &5.99     &5.77  \\
        16  &6.93 & 7.01  &7.13     &6.78     &\textbf{6.58}     &6.60  \\ \hline
    \end{tabular}
\label{table4}
\end{table*}
Explicitly, all experiments were performed with the original loss function ${L_{CD}} + {L_{soft}}({t^2})$ \cite{lang2020samplenet}. Figure \ref{fig8} shows the comparative results of different self-attention blocks on downsampling classification. The method based on local-global attention mechanism almost obtains the best accuracy in all downsampling ratios. This result indicated that convolution with local connectivity and self-attention with global receptive field have complementary properties. However, the resource overhead of local-global attention is not the ideal solution. On the other hand, we find that the output accuracy of LighTN with two-head ($m$ = 2) attention model is just slightly higher than single-head attention model, which is also higher than three-head attention model. For example, set the downsampling ratio to 16, the output accuracy of $QKV (m=1)$, $QKV (m=2)$ and $QKV (m=3)$ are 85.41\%, 85.49\% and 85.37\%, respectively. The reason might be that the increased model complexity improves the difficulty of LighTN training and leads to learning redundant parameters. The complex model structure also brings higher average computation load and storage space cost. For example, in the case of downsampling ratio of 16, comparing $QKV (m=2)$ and $QKV (m=3)$ with $QKV (m=1)$, FLOPs increase by 65.6\% and 131.7\%, Params increases by 1.8\% and 3.6\%, respectively. In summary, this research explores a single-head attention architecture, called self-correlation, designed for downsampling ranges that achieves a competitive balance between performance and resource overhead.

Later, our experiments show that although the parameter matrices provide more learnable parameters for LighTN training, it has little effect on output accuracy for downsampling ranges. The computation and storage cost is optimal after removing all projection matrices. Besides, self-correlation mechanism without projection matrices is more suitable for calculating the attention score because its internal symmetry matrix satisfies the permutation invariance. Finally, the experimental results demonstrate that the proposed self-correlation block has better discrimination ability and minimal resource overhead than single-head dot-production attention.

\textbf{FFN configuration}
Next, we conduct an ablation study on the FFN scaling module with expand-reduce strategy. We take two linear layers, $L(512,m \times 3)$, without expand-reduce ($L = 2, r = 0$) as baseline where the values in $L( \cdot )$ denote the number of nodes in linear layer and $m$ denotes the number of downsampled points. Naturally, the number of nodes in last linear layer cannot be changed. Otherwise, the number of learned points will deviate from the downsampling ratio. Experimental results are depicted in Figure \ref{fig9}. Three layers FFN and ${r_{l2}} = 2$ enable LighTN to get the best output accuracy, and the FlOPs and Params increase by only 0.0087G and 0.0488M on average, respectively. Considering all factors, we set $L=3$ and ${r_{l2}} = 2$ as our default for all models. Besides, if network needs to reduce resource costs further, the setting of $L=3$ and $r=2$ is a good choice.

\textbf{Loss function}
As discussed in section \ref{sec: Final Loss Function}, the proposed loss function of the ${L_{sampling}}$ consists three components: ${L_{CD}}$, ${L_{repl}}$ and ${L_{soft}}$. ${L_{CD}} + {L_{soft}}({t^2})$ is set as the baseline. Next, we test their impact on the performance of LighTN by conducting experiments on ModelNet40. To the project loss ${L_{soft}}$ in classification task, we test two nonlinear function: $T(t) = {t^2}$ used in SampleNet \cite{lang2020samplenet} and $T(t) = {e^t}$. Table \ref{table3} reports the comparison results. For ${L_{soft}}$, exponential function $exp( \cdot )$ is more conducive to convergence of model accuracy. In particular, these experiments indicate that our proposed sampling loss function ${L_{sampling}}$ makes the task network obtain higher accuracy. Besides, more function $T$ of temperature coefficient $t$ are examined on the registration task. We repeat the experiment three times for reasonable testing and used the average rotation error as the output results. Table \ref{table4} shows the rotation error with sampled points of LighTN, which was trained with different $T$. The results in Table \ref{table3} and \ref{table4} indicate that combining ${L_{soft}}({e^2})$ and ${L_{repl}}$ contributes to the best performance obtained by LighTN.

\section{Conclusion}
\label{sec: conclusion}
This paper presents a light-weight Transformer network named LighTN to simplify the point cloud input through task-oriented downsampling. With the help of single-head self-correlation mechanism, the extracted features contain refined global context information, which shows excellent learnable capacity under limited resource overhead. Then, we design a novel sampling loss function consisting of three components guiding the LighTN to focus on edge and prominent point cloud regions while ensuring the simplified point cloud has uniform distribution and adequate coverage. Furthermore, we introduce an expand-reduce strategy to increase the depth of LighTN with lightweight computational and storage costs. Extensive experiments on classification and registration tasks demonstrate LighTN achieves state-of-the-art task-oriented point cloud downsampling.

Certainly, there are some limitations of LighTN. First, the self-correlation mechanism involves massive matrix multiplication operations, which will consume more computation resources than the PointNet-based family of approaches on devices or terminals. Second, point-wise local features have been proven to improve the performance of LighTN in ablation experiments but were not introduced due to resource overhead constraints. In future work, a lower-energy addition replaces the matrix multiplication operation in the self-correlation module, and light-weight convolutional network for performance improvement are the directions of interest.

\bibliographystyle{ieeetr}
\bibliography{LighTN}

\end{document}